\documentclass[letterpaper, 10 pt, conference]{ieeeconf}  

\IEEEoverridecommandlockouts                              

\usepackage{times}

\usepackage{cite} 
\usepackage{multicol}
\usepackage{soul}
\usepackage[bookmarks=true]{hyperref}
\usepackage{graphics} 
\usepackage{epsfig} 
\usepackage{xcolor,colortbl}
\usepackage{color}
\usepackage{amsmath} 
\usepackage{amssymb}  
\usepackage{bm}
\definecolor{sblue}{RGB}{54,122,199}
\hypersetup{
  colorlinks,
  citecolor=sblue,
  linkcolor=sblue,
  urlcolor=black}

\title{\LARGE \bf
Dream to Fly: Model-Based Reinforcement \\ Learning for Vision-Based Drone Flight
}

\author{Angel Romero$^{*}$, Ashwin Shenai$^{*}$, Ismail Geles, Elie Aljalbout, Davide Scaramuzza
\thanks{*These authors contributed equally. 
The authors are with the Robotics and Perception Group, Department of Informatics, University of Zurich, Switzerland (\protect\url{http://rpg.ifi.uzh.ch}).
This work was supported by the European Union’s Horizon Europe Research and Innovation Programme under grant agreement No. 101120732 (AUTOASSESS) and the European Research Council (ERC) under grant agreement No. 864042 (AGILEFLIGHT).
}%
}

\usepackage{eso-pic}
\usepackage{xcolor}
\definecolor{somegray}{rgb}{0.5, 0.5, 0.5}
\newcommand\PaperNotice{%
\AddToShipoutPictureFG*{%
    \AtPageUpperLeft{%
        \put(0,-40){ 
            \parbox{\paperwidth}{%
                \centering
                \color{somegray}\large
                This paper has been accepted for publication at the\\
                IEEE International Conference on Robotics and Automation (ICRA), Vienna 2026. ©IEEE
            }
        }
    }
}
}

\begin{document}
\PaperNotice

\maketitle
\thispagestyle{empty}
\pagestyle{empty}

\begin{abstract}
 Autonomous drone racing has risen as a challenging robotic benchmark for testing the limits of learning, perception, planning, and control.
 Expert human pilots are able to fly a drone through a race track by mapping pixels from a single camera directly to control commands.
 Recent works in autonomous drone racing attempting direct pixel-to-commands control policies have relied on either intermediate representations that simplify the observation space or performed extensive bootstrapping using Imitation Learning (IL).
 %
 %
%
%
%
%
This paper leverages DreamerV3 to train visuomotor policies capable of agile flight through a racetrack using only pixels as observations.
In contrast to model-free methods like PPO or SAC, which are sample-inefficient and struggle in this setting, our approach acquires drone racing skills from pixels.
 %
 %
 Notably, a perception-aware behaviour of actively steering the camera toward texture-rich gate regions emerges without the need of handcrafted reward terms for the viewing direction.
 Our experiments show in both, simulation and real-world flight using a \emph{hardware-in-the-loop} setup with rendered image observations, how the proposed approach can be deployed on real quadrotors at speeds of up to 9 m/s.
 %
 %
 These results advance the state of pixel-based autonomous flight and demonstrate that MBRL offers a promising path for real-world robotics research.
 %
\end{abstract}
\noindent
\textbf{Video:} \href{https://www.youtube.com/watch?v=nctQ2rxZnIc}{https://www.youtube.com/watch?v=nctQ2rxZnIc}

\vspace{-6pt}
\section{INTRODUCTION}
%
In recent years, quadrotors have become a central focus of robotics research, emerging as versatile platforms with untapped potential across multiple domains, including search and rescue, inspection, agriculture, cinematography, delivery, passenger air vehicles, space exploration~\cite{Mohsan23} and drone racing~\cite{hanover2024autonomous}.
%
%
The drone racing domain not only benefits from cutting-edge robotics research~\cite{moon2019challenges} but also pushes the limits of what is possible by challenging these flying machines to outperform the most skilled human pilots, as shown by recent successes against the world's best pilots~\cite{kaufmann2023champion,song2023reachingthelimit}.
%
%
%

%
Traditionally, most autonomous drone racing systems have relied on explicit state estimation integrating data from inertial measurement units (IMUs) and other onboard sensors to maintain stability and optimize performance~\cite{jung2018direct,kaufmann2018DDR, de2021artificial,qin2024time,kaufmann2023champion}.
\begin{figure}[t]
    \centering
    \includegraphics[width=1.0\linewidth]{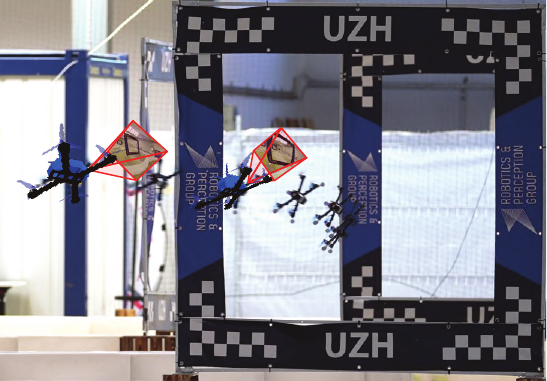}
    \caption{Real-world deployment of our DreamerV3 policy in the Figure~8 track. During training, the agent learns a world model from interactions with the environment. At the same time, the actor-critic policy is trained by sampling the predictions of the world model, also called \emph{imagination}. The rendered images consumed by the network are marked in red.}
    \label{fig:fig1}
    \vspace{-12pt}
\end{figure}
%
%
However, professional human pilots rely solely on visual feedback from a single onboard camera, showcasing a remarkable ability to navigate complex environments purely from visual inputs.
Emulating this human ability to fly based solely on visual information remains a significant challenge for autonomous systems. 
Closing the loop between perception and control -- learning directly from pixels to actions, without the need for explicit state estimation -- remains largely unfulfilled.
While reinforcement learning (RL) has shown promise in various robotic applications, applying it to vision-based tasks introduces unique difficulties in robotics. 
One of the most recent works in this domain \cite{geles2024demonstrating} manages to train a Proximal Policy Optimization (PPO)~\cite{schulman2017proximal} policy to fly a drone through a race track from binary image representations.
In this case, the visual inputs are first preprocessed and distilled into a simpler intermediate representation in the form of a binary mask where only the racing gates are visible.
This intermediate representation reduces the overall observation space complexity and allows the PPO policy to learn the behavior efficiently.
However, this work relies on a simplified observation space: the raw visual input is preprocessed into a binary mask highlighting only the gates. 
This simplification, while reducing the complexity of the observation space and facilitating learning, also discards valuable information. 
For instance, excluding background information can hinder the agent's ability to navigate when no gates are immediately visible, or to obtain the gravity direction from the horizon line, for example.
In \cite{xing2024bootstrapping} the authors present another recent work that tackles quadrotor flight from pixel observations. 
However, it relies on IL bootstrapped from a pre-trained expert policy, which had access to the full state and introduces a strong dependency on privileged information.
Despite their success, one major challenge in robot learning is the need for large amounts of physical interaction data, which can be very expensive and challenging to obtain. 
This challenge is compounded in vision-based control tasks, where the high dimensionality of image observations further increases data requirements. 
%
%

MBRL presents a promising solution to these challenges because it is generally more sample efficient than its model-free counterpart, reducing the need for extensive environment interactions. 
MBRL learns the transition dynamics of the system, known as \textit{world model}, and uses it to predict and optimize future actions. 
It also generalizes well to different tasks as the transition model does not vary vastly across different tasks in the same system description.
However, despite the sample efficiency, this approach typically comes at the cost of longer training times, as the system must simultaneously learn a world model and optimize a control policy.
Additionally, the inherent complexity of model-based RL architectures, with their multiple interdependent components, often makes them more difficult to tune and optimize.

Recent works in MBRL, such as the DreamerV3 architecture~\cite{hafner2023dreamerv3}, have helped mitigate these issues, offering an approach that is more accessible in terms of tuning and optimization. 
%
Although DreamerV3 has shown strong results in controlled settings, its adoption in real-world robotic systems, particularly for pixel-based tasks, remains limited.
We build on DreamerV3 to demonstrate its effectiveness in vision-based robotics tasks that require egocentric perception and agile control, such as drone racing. Our contributions are as follows:
%

%
%
%
%
%
%
\begin{itemize}
\item \textbf{Pixel-to-command MBRL system for quadrotor flight:} We train quadrotor policies from scratch, without relying on intermediate representations or bootstrapping from imitation learning, which were necessary in previous works.
\item \textbf{Emergent perception-aware behaviour:} the drone's camera view is naturally guided towards feature-rich areas such as the next gates. 
This behavior arises directly from our end-to-end optimization from pixels to commands, without requiring handcrafted reward terms that were common in previous methods. 
\item \textbf{Real-world policy transfer:} We validate our learned policies by deploying them on a physical quadrotor with a \emph{hardware-in-the-loop} (HIL) setup.
\end{itemize}
Our approach demonstrates that the learned policy effectively controls real-world dynamics directly from rendered pixel observations, and further validates the applicability and potential of MBRL for real-world mobile robotic tasks.
%
%
%
%

\section{RELATED WORK}
\subsection{Reinforcement Learning from Pixels}
In recent years, RL algorithms have achieved incredible feats in simple environments learning directly from pixels, such as arcade games, used as sample efficient simulators to benchmark the capabilities of AI agents~\cite{bellemare2013arcade}.
This success in RL has been repeatedly shown in environments in growing complexity, 
such as beating human performance in Atari games ~\cite{mnih2015humanlevelcontrol} or in more complex games such as Go~\cite{silver2016go}.
However, these successes have been primarily in environments where the action space is discrete, and not in continuous action spaces.
This gap is addressed by the DeepMind Control Suite~\cite{tassa2018controlsuite}, which presents a diverse set of environments with different observation spaces and different action modalities --- both continuous and discrete.
However, learning directly from pixels in these scenarios has been shown to be less sample efficient, yielding suboptimal performance when compared to state-based learning, even in simulation environments.
Recent efforts have tried to improve this by building on top of intermediate visual representations~\cite{yarats2021sampleefficiency, aljalbout2021learning,hafner2019learninglatent,laskin2020curl,geles2024demonstrating}.
For example, previous work explored different kinds of state representations via various supervised and self-supervised learning methods, such as auto-encoders~\cite{yarats2021sampleefficiency}, future predictions~\cite{hafner2019learninglatent}, contrastive unsupervised representations~\cite{laskin2020curl} or semantic segmentation~\cite{geles2024demonstrating}, with the objective of reducing the performance difference between state-based and pixel-based RL.

In order to address the challenge of learning complex behaviors efficiently directly from pixels, some methods have focused on data augmentation~\cite{laskin2020rlwithaugmenteddata, yarats2022visualcontrol}.
Policies that map camera observations directly to commands --- also called sensorimotor or visuomotor policies --- are mostly learned through extensive data simulation, making use of domain randomization techniques or incorporating additional privileged information such as joint angles ~\cite{rusu2017sim2real, tobin2017domainrandomization, levine2016end2end}
For more complicated tasks, the robotics community has resorted to the usage of expert demonstrations, marking a shift towards imitation learning approaches, contrary to the learning from scratch using RL paradigm. This is showcased by recent work, most notably ~\cite{fu2024mobilealoha}.
\subsection{Model-based RL for Real-World Robotics}
\label{sec:related_mbrl}
As mentioned above, there are many approaches that have attempted to train policies using pixels as observations, relying in simplifying assumptions.
However, the landscape of model-based RL (MBRL) methods applied to real-world robotics remains relatively limited, and existing research primarily focuses on simulation environments.
%
%
Within the context of real-world robotic applications, prior MBRL work often relies on state estimations rather than raw pixel observations.
Specifically, there are some works tackling the problem of flying a quadcopter using model-based RL, for high-level control~\cite{becker2020learning} or low-level controller learning~\cite{lambert2019low}. However, these works have been using state estimation as observations, and not directly pixels as observations.
When speaking about works that use MBRL from pixels and that deploy in real-world robots, the number of works is even lower.
One of the early works in this direction is \cite{wahlstrom2015pixels}, where a learned world model of the dynamics is built from pixels, and is then used by an MPC planner to get the optimal policy, which outputs torque commands for a real-world manipulator. 
Another interesting work that applies DreamerV3 to a real-world robotic system is \cite{bi2024sample}, where the table top labyrinth game is solved in the real world by using top-view fixed camera in conjunction with the measurement of the table angle. However, this setup does not fall under mobile robotics, as the observations are not egocentric but rather taken from a fixed viewpoint with most of the environment remaining unchanged in the field of view.
One of the most salient works that applies DreamerV3 \cite{hafner2023dreamerv3} to real-world robotics is~\cite{wu2022daydreamer}, where different policies are trained from images, depth and state, for different tasks. 
These include pick-and-place manipulation utilizing a fusion of RGB images, depth data, and proprioceptive sensing, as well as quadruped locomotion trained solely on proprioceptive sensor data.
Notably, the only instance of direct learning from pixels to commands involves a 2D navigation task with a Sphero robot, a relatively simple scenario in terms of dynamics and environmental complexity.

\begin{figure*}[ht]
    \centering
    \includegraphics[width=0.9\linewidth]{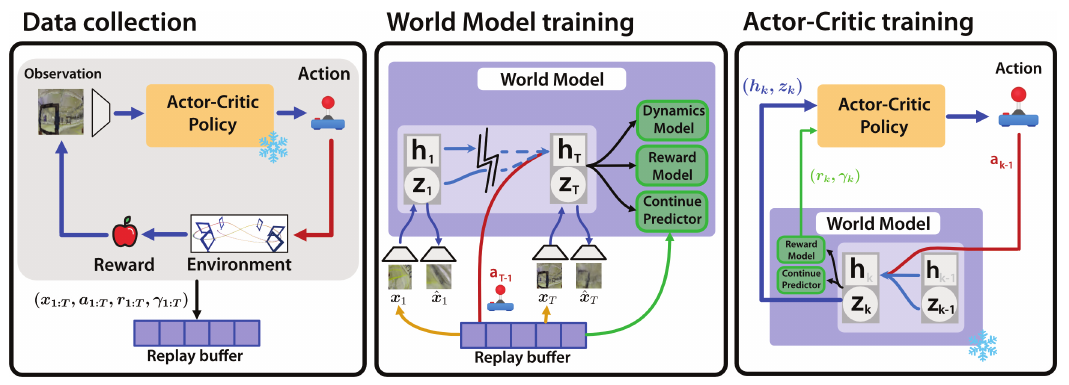}
    \caption{The process begins with data collection in the simulation environment using the current policy, storing experiences in a replay buffer. This buffer is used to train the world model components: the encoder, decoder, RSSM, dynamics model, reward model, and continue predictor (Section \ref{sec:dreamerv3}). Subsequently, an actor-critic policy is trained within the learned world model to maximize expected (imagined) returns. This updated policy is then used to collect new data, restarting the loop.}
    \label{fig:architecture}
    \vspace{-14pt}
\end{figure*}
\section{METHODOLOGY}
\subsection{Problem Statement}
%
Formally, we seek a policy $\pi(x)$ that maps raw visual observations $x$ directly to control commands $a$, minimizing the time required to complete the course.
We address this challenge using reinforcement learning (RL), framing the problem within the Partially Observable Markov Decision Process (POMDP) framework.
%
A POMDP is defined by the tuple $(\mathcal{S}, \mathcal{A}, \mathcal{X}, P, R, \gamma, \Omega)$, where $\mathcal{S}$ is the set of latent states, $\mathcal{A}$ is the set of actions, $\mathcal{X}$ is the set of observations, $P(s_{k+1} \mid s_k, a_k)$ is the state transition probability, $R(s_k, a_k)$ is the reward function, $\gamma \in [0,1)$ is the discount factor, and $\Omega(x_k \mid s_k)$ is the observation model.
%
%
In our vision-based drone racing task, the latent state $s_k$ is not observed directly; instead, the agent receives the RGB image observation $x_k \in \mathcal{X}$ rendered at time step $k$.
%
%
%
The action $a_k$ consists of the collective thrust and body rates applied to the quadrotor. 
The reward function is designed to incentivize fast and collision-free navigation through the gates (detailed in Section \ref{sec:reward_design})
The objective in a POMDP is to find an optimal policy $\pi_\theta^*: \mathcal{X} \rightarrow \mathcal{A}$ that maximizes the expected cumulative discounted reward by optimizing the parameters $\theta$ of a neural network,
\begin{equation}
\pi_{\theta}^* = \arg\max_{\pi} \mathbf{E}\left[\sum_{k=0}^{\infty} \gamma^k r_k \right]. \notag
\end{equation}
%
%
This optimization considers both immediate and future rewards, with future rewards discounted by $\gamma$.
%
%

\subsection{Observation and Action Spaces}
The RL framework learns a visuomotor, end-to-end policy that directly maps raw RGB images as observations to control inputs.
These images provide a rich and high-dimensional representation of the environment, enabling the agent to infer its state and surroundings without relying on explicitly estimated states.
\subsubsection{Observation Space}

The observation space consists of RGB images captured at each time step, denoted as $\mathbf{x}_k \in \mathbb{R}^{H \times W \times 3}$, where $H$ and $W$ represent the image height and width, respectively, and 3 corresponds to the RGB color channels. 
These images are normalized to the range $[0, 1]$ by dividing pixel values by 255.
Unlike approaches that rely on explicit state estimation (e.g., position, velocity, or attitude), this image-based approach directly leverages rich visual information to guide the policy.
\subsubsection{Action Space}
At each time step $k$, the policy outputs a four-dimensional action vector $\mathbf{a}_k = [c, \omega_x, \omega_y, \omega_z]$, where $c$ represents the mass-normalized collective thrust, and $\omega_x, \omega_y, \omega_z$ are the body rate setpoints in the drone's body frame. 
These actions are expressed in the Collective Thrust and Body Rates (CTBR) format, a control interface commonly used by professional drone pilots~
\cite{kaufmann2023champion, pfeiffer2021human}. 
This format directly commands the low-level actuation of the drone, bypassing the need for intermediate, high-level abstractions such as position or velocity commands. 
To ensure bounded and physically realizable control actions, the action space is constrained to $\mathcal{A} = [-1, 1]^4$, which are then mapped to the actual collective thrust and body rates limits.
This constraint is enforced by applying a hyperbolic tangent (tanh) activation function to the output of the actor policy network.
\subsection{Reward Function}
\label{sec:reward_design}

The racing track is defined as a sequence of linearly connected waypoints placed at the center of the gates.
The reward function is designed to incentivize progress along this track while penalizing undesirable behaviors. The reward at time step $k$, $r(k)$, is defined as
%
\begin{equation}
    r_k =
    \begin{cases}
        r_{collision}, & \text{if collision;} \\
        r_{passed}, & \text{if gate passed;} \\
        \begin{aligned}[b]
            b_1(&\Vert \mathbf{g}_k - \mathbf{p}_{k-1} \Vert - \Vert \mathbf{g}_k - \mathbf{p}_k \Vert) \\ &- b_2 \| \boldsymbol{\omega}_k \|, 
        \end{aligned}& \text{otherwise,}
    \end{cases}
    \label{eq:gate_progress}
\end{equation}
%
where $r_k$ is the reward at timestep $k$, $\mathbf{g}_k$ is the center of the target gate, $\mathbf{p}_k$ and $\mathbf{p}_{k-1}$ are the drone's positions at the current and previous time steps, respectively, and $\boldsymbol{\omega}_k$ is the body rate vector. 
The primary component of the reward, $b_1(\Vert \mathbf{g}_k - \mathbf{p}_{k-1} \Vert - \Vert \mathbf{g}_k - \mathbf{p}_k \Vert)$, is called the progress term, as it encourages the drone to move closer to the target gate. 
The term $b_2 \| \boldsymbol{\omega}_k \|$ penalizes excessive body rates. We use the coefficients $b_1 = 1.0, b_2 = 0.01$.
This formulation directly rewards progress toward the gate center. 
Importantly, deviations from the exact path defined by the waypoints are not penalized, allowing the agent to discover potentially more efficient trajectories.

In addition to the continuous reward component, discrete rewards are provided for specific events, e.g., $r_{collision}=-4.0$ and $r_{passed}=+10.0$.
%
%
%
%
\subsection{Model-based Reinforcement Learning: DreamerV3}
\label{sec:dreamerv3}
In this section, we briefly describe the key aspects of the DreamerV3 algorithm. For more details, we refer the reader to the original paper~\cite{hafner2023dreamerv3}.
DreamerV3 is an off-policy, model-based reinforcement learning algorithm. 
The algorithm is structured in two main blocks: the \textbf{world model}, and the \textbf{actor-critic policy}.
These two blocks are trained in an alternating fashion using an experience replay buffer while the agent interacts with the environment.
A high level depiction of the training process is described in Fig.~\ref{fig:architecture}.
\subsubsection{World Model training}
Our approach employs a world model that captures the state transition dynamics in a compact latent space. 
Specifically, we consider a latent state \(\bm{s}_k = (\bm{h}_k, \bm{z}_k)\) and model the state transition probability \(P(\bm{s}_{k+1} \mid \bm{s}_k, \bm{a}_k)\). 
By encoding high-dimensional sensory inputs into low-dimensional representations, the world model enables predicting future latent states and rewards based on the agent's actions.
The world model is implemented as a Recurrent State Space Model (RSSM)~\cite{hafner2019learninglatent}, and consists of the following components:

\begin{itemize}
    \item \textbf{Encoder:} An encoder network maps raw observations \(\bm{x}_k\) into a stochastic latent representation \(\bm{z}_k\). This provides a compressed representation of the sensory observations.
    \begin{align}
       \bm{z}_k \sim &\ q_\phi(\bm{z}_k \mid \bm{h}_k,\bm{x}_k).
    \end{align}
    \item \textbf{Recurrent Sequence Model:} A recurrent sequence model, parameterized by the recurrent state \(\bm{h}_k\), predicts the evolution of the latent representation. Given the previous latent state $[h_{k-1}, z_{k-1}]$ and the action \(\bm{a}_{k-1}\), it predicts the current recurrent state \(\bm{h}_k\).
    \begin{align}
       \bm{h}_k = &\ f_\phi(\bm{h}_{k-1},\bm{z}_{k-1},\bm{a}_{k-1}).
    \end{align}
    \item \textbf{Dynamics, Reward and Continue Prediction:} The dynamics predictor predicts the stochastic state $\hat{\bm{z}}_k$ given the recurrent state $\bm{h}_k$.
    The reward and continue predictors are conditioned on the latent state \(\bm{s}_k = (\bm{h}_k, \bm{z}_k)\), and predict the immediate reward \(r_k\) and the episode continuation flag \(c_k \in \{0,1\}\), indicating whether the episode terminates or continues.
    \begin{align}
    & \text{Dynamics predictor:}   && \hat{\bm{z}}_k      &\sim&& p_\phi(\hat{\bm{z}}_k &\mid \bm{h}_k) \\
    & \text{Reward predictor:}     && \hat{\bm{r}}_k      &\sim&& p_\phi(\hat{\bm{r}}_k &\mid \bm{h}_k,\bm{z}_k) \\
    & \text{Continue predictor:}   && \hat{\bm{c}}_k      &\sim&& p_\phi(\hat{\bm{c}}_k &\mid \bm{h}_k,\bm{z}_k).
    \end{align}
    \item \textbf{Decoder:} A decoder reconstructs the original observations from the latent representations. This reconstruction loss ensures that the latent variables \(\bm{z}_k\) retain essential information from the environment.
    \begin{align}
    \hat{\bm{x}}_k \sim &\ p_\phi(\hat{\bm{x}}_k \mid \bm{h}_k,\bm{z}_k).
    \end{align}
\end{itemize}

The encoder and decoder use convolutional neural networks (CNNs) for image inputs and multi-layer perceptrons (MLPs) for vector inputs. 
These models are trained by minimizing different losses: the prediction loss $\mathcal{L}_{pred}(\phi)$, which trains the decoder, the reward and the continue flags; the dynamics loss $\mathcal{L}_{dyn}(\phi)$, which trains the sequence model; and the representation loss $\mathcal{L}_{rep}(\phi)$, aims to make the representations more predictable.
For more details about these losses, we refer the reader to \cite{hafner2023dreamerv3}.
%
%

The overall world model objective is a linear combination of the above defined losses:
\begin{small}
\begin{align*}
    \mathcal{L}(\phi) = \mathbf{E}_{q_\phi}\left[\sum_{k=1}^T \beta_{pred} \mathcal{L}_{pred}(\phi) + \beta_{dyn} \mathcal{L}_{dyn}(\phi) +
    \beta_{rep} \mathcal{L}_{rep}(\phi) \right],
\end{align*}
\end{small}
where $\beta_{pred} = 1, \beta_{dyn} = 1, \beta_{rep} = 0.1$ and $T$ is the batch sequence length. The world model is trained by randomly sampling $T$-length snippets of inputs $\bm{x}_{1:T}$, actions $\bm{a_}{1:T}$, rewards $r_{1:T}$ and continuation flags $c_{1:T}$ from episodes in the replay buffer.

By jointly training these components, the RSSM-based world model learns a compact, predictive representation of the environment, and a latent dynamics model, supporting more efficient decision-making and planning in latent space.



%
%
\subsubsection{Actor-Critic Training}
Our approach employs an actor-critic architecture trained using imagined trajectories generated by a learned world model (see Fig. \ref{fig:architecture}). 
This allows the agent to learn complex behaviors without requiring extensive real-world interactions. 
Given a starting state representation, the actor generates an imagined trajectory consisting of model states $(\mathbf{h}_{1:T}, \mathbf{z}_{1:T})$, actions $\mathbf{a}_{1:T}$, rewards $r_{1:T}$, and continuation flags $c_{1:T}$.
The critic then learns to evaluate the quality of these imagined trajectories by predicting the distribution of bootstrapped $\lambda$-returns. This bootstrapping allows the critic to estimate long-term returns even within the limited prediction horizon of the world model.
The actor learns to maximize these $\lambda$-returns while exploring through an entropy regularizer. To ensure robust exploration across diverse reward scales and frequencies, the returns are normalized to approximately lie in $[0,1]$.
To optimize the policy, DreamerV3 uses the REINFORCE estimator combining unbiased but high-variance policy gradients with a stop-gradient operation on the value targets.

For further details on the actor-critic training and loss functions, we refer the reader to \cite{hafner2023dreamerv3}. In our case, the prediction horizon for the imagined trajectories is set to $T = 16$. This horizon balances computational cost and the ability to capture longer-term dependencies.
%
\begin{figure*}[ht]
    \centering
    \includegraphics[width=0.92\linewidth]{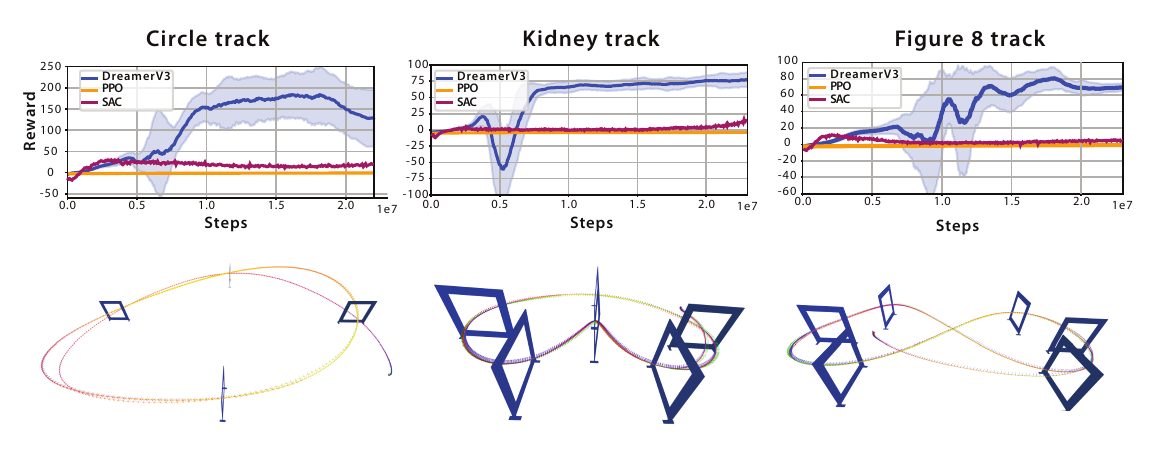}
    \caption{Reward evolution by number of steps for three different tracks: Circle track, Kidney Track and Figure 8 track. The training performance of DreamerV3 is shown in blue, for PPO in orange and for SAC in red. We show that none PPO nor SAC are able to achieve any considerable training in 20 million environment interactions, while DreamerV3 is able to train to convergence for the three tracks.}
    \label{fig:reward_evolution}
    \vspace{-8pt}
\end{figure*}
\begin{figure}[ht]
    \centering
    \includegraphics[width=0.95\linewidth]{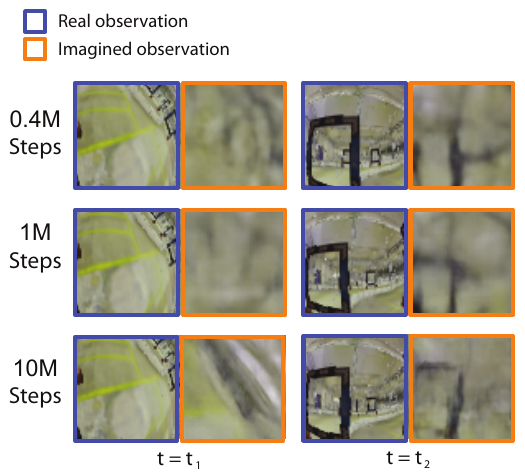}
    \caption{Comparison of real observations and imagined observations for the Figure 8 track. Imagined observations are observations that are reconstructed by the world model. The figure shows the reconstructed observations after 0.4M steps (early stage training), 1M timesteps (mid stage training), and after 10M steps (training convergence). One can observe how the reconstruction gets better and better as the training evolves.}
    \label{fig:imag}
    \vspace{-16pt}
\end{figure}
\section{Simulation Experiments}
\label{sec:sim_exp}
We implemented DreamerV3 in PyTorch, leveraging the dreamerv3-torch open-source implementation and building on top of the stable-baselines3 library \cite{stable-baselines3}. 
%
Our experiments were conducted within a high-fidelity simulation environment combining Flightmare \cite{yunlong2020flightmare} and Agilicious \cite{foehn2022agilicious} for realistic quadrotor dynamics and track generation. 
To enable fast, real-time rendering and direct access to the image feed during training, we integrated the Habitat simulator \cite{habitat19iccv, puig2023habitat3}, enabling training with the renderer in the loop at several thousand frames per second. 
This same training environment has been successfully employed in prior research \cite{xing2024bootstrapping}.

Our method's performance is benchmarked against two model-free baselines: Proximal Policy Optimization (PPO) \cite{song2021autonomous, song2023reachingthelimit, kaufmann2023champion, geles2024demonstrating} and Soft Actor Critic (SAC)~\cite{haarnoja2018soft}. 
Both the PPO and SAC baselines used a CNN followed by four-layer Multi-Layer Perceptrons (MLPs) with 768 neurons per layer for both the actor and critic networks.
The physical parameters of the quadrotor used were consistent across all experiments: mass $m = 0.6$ kg, diagonal inertia matrix $J = \text{diag}([0.002410, 0.001800, 0.003759])$ kg m$^2$, rotor torque constant $\kappa = 0.022$, and arm length $0.14$ m. The maximum rotor thrust was limited to $4.0$ N, resulting in a thrust-to-weight ratio of $2.7$.
The dynamics, platform and parameters are the same one as in ~\cite{geles2024demonstrating}.

For DreamerV3, we adopted the hyperparameters detailed in \cite{hafner2023dreamerv3} and used the \textit{Large (L)} network configuration described in Appendix B of \cite{hafner2020dreamer}. 
This configuration consists of four-layer MLPs with 768 units for the decoder, predictors, actor, and critic networks, and 2048 recurrent units for the world model's recurrent component.

All experiments were performed on a single Quadro RTX 8000. Input images from the simulated camera were resized to $64 \times 64$ RGB pixels before being fed into the DreamerV3 agent. These input images are visualized in Figure \ref{fig:imag}.
\subsection{Results}
We conducted training experiments on our agent using three distinct reinforcement learning algorithms: DreamerV3, PPO and SAC. 
These experiments were carried out across three challenging drone racing tracks: Circle, Kidney, and Figure 8.
The results of these experiments are visualized in Figure \ref{fig:reward_evolution}. 
The top row of the figure presents the overall reward evolution for each track, comparing the performance of both DreamerV3, PPO and SAC. 
%
%
Each run was conducted with 5 different random seeds.
The resulting plots depict the average reward across these 5 runs, represented by a solid center line, and the shaded area surrounding the line indicates the standard deviation.
Our findings indicate that DreamerV3, by leveraging a learned world model, effectively learns directly from pixel observations. 
This enables the agent to acquire policies that are capable of racing through each track.
In contrast, PPO, and similarly SAC, which rely solely on direct interaction with the environment, struggle to converge to high-reward solutions.
Furthermore, the model-free policies fail to execute any meaningful flight maneuvers, resulting in consistently low reward values.
An important observation is that our system is trained using curriculum learning, where the parameter $b_2$ in Eq. \eqref{eq:gate_progress} is initially set to $0.0$ and gradually increases once the total reward surpasses $50.0$. 
This explains the oscillatory reward values observed in Fig. \ref{fig:reward_evolution}, as the system adjusts to the changing rewards.
Our experimental results align with the findings presented in \cite{xing2024bootstrapping}, where the authors encountered similar challenges in training end-to-end PPO policies directly from pixel observations. 
To overcome these difficulties, they adopted a hybrid approach, combining imitation learning with subsequent reinforcement learning fine-tuning. In contrast, using a model-based approach, we are able to completely avoid such additional efforts.
We note, however, that this comes at the cost of a significantly longer training time than the baselines, reaching 240 hours.
%
%

%
%

%
As illustrated in Figure \ref{fig:architecture}, the DreamerV3 architecture incorporates an observation reconstruction module. This component ensures that the latent state captures sufficient information to accurately represent the observed sensory input.
%
%
%
Figure \ref{fig:imag} presents a comparative analysis of real observations and their corresponding reconstructions for the Figure 8 track at two distinct time steps, $t_1$
and $t_2$. The reconstructions are sampled at three different training stages: early training (0.4 million steps), mid-training (1 million steps), and late-stage training (10 million steps).
A visual inspection of the figure reveals a progressive improvement in reconstruction quality as training progresses. Notably, the fine details of the environment, such as the yellow floor lines at time step $t_1$, become identifiable only in the late-stage reconstructions. Similarly, the shape of the gate at time step $t_2$ is accurately reconstructed by the model only at late-stage training. 
\subsection{Perception-aware emergent behaviour}
\label{sec:emergent}
\begin{figure}
    \centering
    \includegraphics[width=0.85\linewidth]{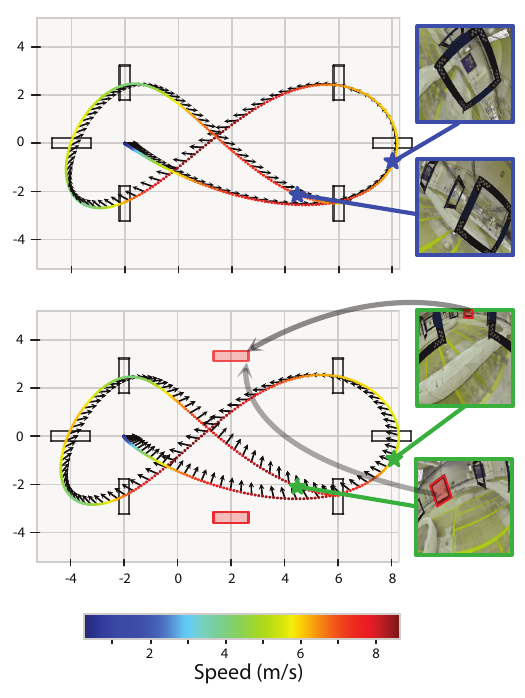}
    \caption{\textbf{Ablation study on the perception aware behaviour of our policies.} \textbf{Top:} DreamerV3 policy trained on pixel observations in an environment where the only rendered gates are the actual gates. As indicated by the black arrows (representing camera direction), the platform predominantly focuses on the next gate. \textbf{Bottom:} We introduce two additional gates to the rendering engine (marked in red color). These gates are not required to be traversed, and are not part of the objective, but serve as a valuable source of information for platform localization. Consequently, the policy's behavior shifts, and the platform now distributes its camera view more evenly across both the actual and the extra gates.}
    \label{fig:emergent}
   \vspace{-12pt}
\end{figure}
An additional interesting observation is that after training, the agent exhibits a consistent tendency to orient the camera towards the gates, a behavior that naturally emerged during training rather than being explicitly incentivized by the reward function, as it is generally done in previous works \cite{geles2024demonstrating, xing2024bootstrapping}.
This emergent behavior can be attributed to the fact that we are optimizing end-to-end from pixels to commands, therefore allowing for closing the action-perception loop.
%
%
As depicted in Figure \ref{fig:imag}, the gates remain visually clear throughout the entire track, while the background details become increasingly blurred due to the downsampling of the input image to $64 \times 64$ pixels. 
This suggests that the policy strategically prioritizes focusing on the information-rich gates, which are essential for successful navigation.
To reinforce this hypothesis, we have conducted an additional ablation experiment where we place two additional gates in the periphery of the Figure 8 track.
These gates are there only in the rendering, but do not need to be passed through.
Fig. \ref{fig:emergent} shows the same policy trained with and without these additional gates.
In the top part of Fig. \ref{fig:emergent}, we show the policy trained without the extra gates.
As indicated by the black arrows (representing camera view direction), the platform predominantly focuses on the following gates. 
In the bottom part of the figure, we show the policy trained with the two additional gates to the rendering engine (marked in red color).
As it can be seen, the policy’s behavior changes, and the platform now distributes its camera view more evenly across both the actual and the extra gates.
In the supplementary video we show the first person rendered view of the two policies depicted in Fig. \ref{fig:emergent} deployed in a simulated environment.

%
%
%
\section{Real-world Experiments}
\label{sec:exp}
\subsection{Setup}
The software setup is identical to the one used in the simulation experiments, explained in Section \ref{sec:sim_exp}. 
Regarding the hardware, we use a modification of the \emph{Agilicious} platform \cite{foehn2022agilicious} for the real-world deployment.
%
We have replaced the onboard computer with an RF receiver, which is connected directly to the \textit{Betaflight} flight controller and takes care of parsing the collective thrust and bodyrate commands from the offboard computer.
Our hardware setup is the same setup as in~\cite{geles2024demonstrating}, similar to the one used by professional drone racing pilots.
%
%
%
%
%
For the deployment in the real world, we use a hardware-in-the-loop (HIL) setup, where the images are rendered using the habitat simulator and fed into the network, and the commands produced are sent directly to the real drone platform.
This way, we have the real world dynamics in the loop, which allows us to asses the sim-to-real gap and our policy performance when tested on the real system.
\subsection{Results}
\begin{figure}
    \centering
    \includegraphics[width=0.85\linewidth]{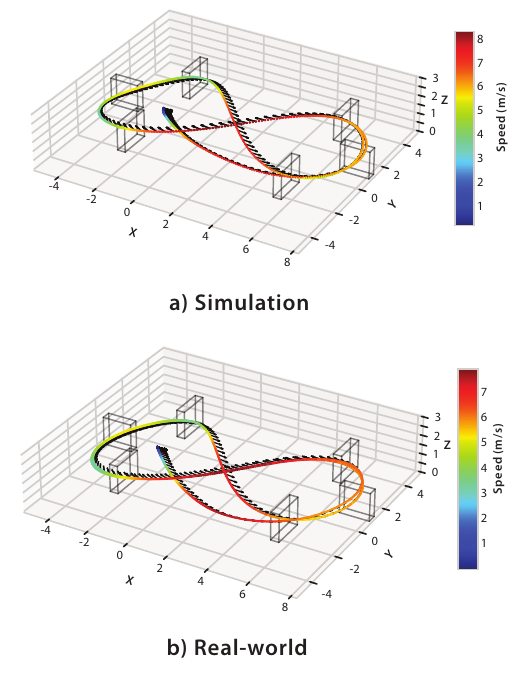}
    \caption{Real-world deployment of the trained policy for the Figure 8 track. We show the deployment in simulation (top) and in the real-world (bottom). By looking at the 3D trajectories and the speed profile, we note that our policies transfer and result in a small sim-to-real gap.}
    \label{fig:real_world}
    \vspace{-10pt}
\end{figure}

We demonstrate the efficacy of our policies by deploying them in the real world on the Figure 8 track. 
This experiment is shown in the supplementary material video at \href{https://www.youtube.com/watch?v=nctQ2rxZnIc}{https://www.youtube.com/watch?v=nctQ2rxZnIc}.
Figure \ref{fig:real_world} presents a comparative analysis of the simulated and real-world trajectories and speed profiles for this track. 
A strong similarity is observed between the two, indicating a minimal sim-to-real gap in our dynamic model. 
Moreover, the camera axis, visualized by black arrows in Figure \ref{fig:real_world}, aligns closely between the simulated and real-world scenarios.
As already mentioned in Section \ref{sec:emergent}, both the simulated and deployed policies show the same perception-aware behavior: they keep the camera view aligned with the next gate, even if there are no reward terms guiding or incentivizing it.
To the best of our knowledge, our HIL demonstration marks the first RL approach to learn drone racing directly from pixel inputs to control commands, effectively closing the loop between perception and action.
The supplementary video shows the deployment of our system in the real world.

%
%
%
%

\section{Conclusion}
This paper presented a MBRL approach using DreamerV3 to train end-to-end visuomotor policies for agile drone flight. 
Our method learns directly from raw pixel inputs, removing the need for intermediate representations or imitation learning bootstrapping. 
Our experiments demonstrated that this approach is more sample-efficient than model-free baselines like PPO and SAC and results in emergent perception-aware behaviors without explicit reward shaping.
We validated the learned policies in simulation and on a physical quadrotor using a hardware-in-the-loop (HIL) setup.
While this confirms the policy's effectiveness on the real quadrotor dynamics, additional challenges remain to transfer to real camera pixel observations. 
Additionally, the training process is computationally intensive, requiring approximately 240 hours to converge. 
Future work should focus on using photorealistic simulators or exploring other observation modalities that allow for deployment directly real pixels, and on improving the computational efficiency of the world model. 
Overall, our work is a promising step towards applying pixel-to-command MBRL on real-world robotic systems.

%

\vspace{-10pt}

\bibliographystyle{IEEEtran}
\bibliography{references}



\end{document}